\documentclass[twoside,11pt]{article}

\usepackage{jmlr2e}

\jmlrheading{1}{}{}{}{}{}

\firstpageno{1}

\usepackage{algorithm}
\usepackage[noend]{algpseudocode}

\usepackage{amsmath}
\usepackage{bbm}
\usepackage{bm}
\usepackage{color}
\usepackage{dirtytalk}
\usepackage{dsfont}
\usepackage{enumerate}
\usepackage{enumitem}
\usepackage{fullpage}
\usepackage{listings}
\usepackage{mathtools}
\usepackage{natbib}
\usepackage{times}
\usepackage{xspace}
\usepackage{hyperref}
\usepackage{url}
\usepackage{cancel}

\usepackage{pifont}

\usepackage[dvipsnames]{xcolor}

\newcommand{\graphfamily}{\mathcal{F}}

\newcommand{\graph}[0]{\boldsymbol{\Gamma}}

\newcommand{\wt}[0]{\mathbf{w}}

\newcommand{\data}[0]{\mathfrak{D}}
\newcommand{\metric}[0]{\boldsymbol{\mu}}

\newcommand{\loss}[0]{\ell}
\newcommand{\losses}[0]{\{\loss_i\}}
\newcommand{\grad}[0]{\mathbf{g}}
\newcommand{\gradi}[0]{\grad_j}
\newcommand{\grads}[0]{\{\gradi\}}

\newcommand{\batch}[0]{B}
\newcommand{\gbatch}[0]{\{\batch_g\}}

\newcommand{\fullEVE}[0]{Ever Evolving Evaluator (EV3)}

\usepackage{natbib}
\usepackage{graphicx}
\usepackage{enumitem}

\usepackage{xspace}
\newcommand{\eg}{{\em e.g.}\xspace}
\newcommand{\ie}{{\em i.e.}\xspace}

\newlist{rqitems}{enumerate}{1}
\setlist[rqitems, 1]{label*={\bf RQ\arabic*}:, ref={\bf RQ\arabic*}, leftmargin=12mm, rightmargin=0pt}

\begin{document}

\title{Ever Evolving Evaluator (EV3): Towards Flexible and Reliable Meta-Optimization for Knowledge Distillation
}

\author{\name Li Ding\thanks{Work done during an internship at Google.} \email liding@umass.edu \\
       \addr University of Massachusetts Amherst \\
       Amherst, MA 01002, USA
        \AND
       \name Masrour Zoghi \email mzoghi@google.com \\
       \name Guy Tennenholtz \email guytenn@google.com \\
       \name Maryam Karimzadehgan \email maryamk@google.com \\
       \addr Google Research\\
       Mountain View, CA 94043, USA
}


\maketitle

\begin{abstract}
We introduce EV3, a novel meta-optimization framework designed to efficiently train scalable machine learning models through an intuitive explore-assess-adapt protocol.  
In each iteration of EV3, we explore various model parameter updates, assess them using pertinent evaluation methods, and then adapt the model based on the optimal updates and previous progress history.  
EV3 offers substantial flexibility without imposing stringent constraints like differentiability on the key objectives relevant to the tasks of interest, allowing for exploratory updates with intentionally-biased gradients and through a diversity of losses and optimizers. 
Additionally, the assessment phase provides reliable safety controls to ensure robust generalization, and can dynamically prioritize tasks in scenarios with multiple objectives.
With inspiration drawn from evolutionary algorithms, meta-learning, and neural architecture search, we investigate an application of EV3 to knowledge distillation. 
Our experimental results illustrate EV3's capability to safely explore the modeling landscape, while hinting at its potential applicability across numerous domains due to its inherent flexibility and adaptability. 
Finally, we provide a JAX implementation of EV3, along with source code for experiments, available at: \url{https://github.com/google-research/google-research/tree/master/ev3}.
\end{abstract}


\section{Introduction}

This work proposes Ever Evolving Evaluator (EV3), a versatile meta-optimization framework for training scalable machine learning models.  
We introduce a new class of optimizers that incorporate an explore-assess-adapt protocol, carrying out the following steps in each iteration of the training loop:
\begin{enumerate}[noitemsep,topsep=0pt,leftmargin=12pt]
    \item {\bf Explore} the model parameter space by proposing a variety of updates to model parameters, \eg, by doing gradient-descent using different loss functions and optimizers.
    \item {\bf Assess} the proposals using one or more evaluation metrics that are relevant to the tasks of interest but are often non-differentiable and cannot be directly used in optimization. The assessments are performed on unbiased data to choose the best updates. 
    \item {\bf Adapt} model parameters and/or topologies based on the results of statistical significance tests, which compare the performance of the best updates to the top-performing models previously identified by EV3. This adaptation also considers factors such as the history of progress.
\end{enumerate}

EV3 has a number of desirable properties that merit elaboration:  
\begin{itemize}[noitemsep,topsep=0pt,leftmargin=17pt]
    \item[\ding{43}] The explore step can be used to perform hyperparameter tuning for losses, optimizers, and schedules.
    \item[\ding{43}] Given sufficient training data or a sufficiently complex learning task \citep{feldman2019open}, the assess and the adapt steps of EV3 could minimize overfitting by using a validation set or unseen online data.
    \item[\ding{43}] The assess and the adapt steps do not impose a differentiability constraint on the evaluation methods:  
    indeed, all that is required is the ability to compare models (\eg, using human feedback on preferences).
    \item[\ding{43}] In the adapt step, updates generated by the previous steps can be rejected, thus eliminating the necessity for every proposed update to result in an improved model. This has a few important corollaries:
    \begin{itemize}[noitemsep,topsep=0pt,leftmargin=16pt]
        \item[\ding{226}] The gradients used to propose updates need not be unbiased, allowing for exploration in the sampling of gradient batches: \eg, boosted sampling \citep{schapire2012boosting} or keeping only particular types of samples  \citep{hadsell2020embracing,soviany2022curriculum}.  
        \item[\ding{226}] Distinct losses and optimizers can be used to generate updates when there is no clear winner.  
        \item[\ding{226}] More generally, the explore step can utilize multiple sources of knowledge or supervision (\eg, different teacher models in knowledge distillation).  
    \end{itemize}
    \item[\ding{43}] In the multi-objective setting, the adapt step can prioritize certain evaluation measures and deprioritize others in addition to updating the model.
\end{itemize}

The particular design of EV3 draws inspiration from multiple domains including evolutionary optimization \citep{ding2021optimizing}, meta-learning \citep{flennerhag2022bootstrapped}, preference-based reinforcement learning \citep{wirth2017survey}, neural architecture search \citep{elsken2019neural}, and gradient-free optimization methods such as Bayesian optimization \citep{shahriari2015taking}.

In this paper, we assess the utility of EV3 to a particular application domain, knowledge distillation (KD), in order to highlight its ability to perform optimization easily and safely. 
We also provide an open-source implementation of EV3 in JAX~\citep{jax2018github} to facilitate its application in more domains.

\section{Methods}

In this section, we first introduce EV3 as a generic meta-optimization framework for ML models, then focus on an implementation of EV3 for training deep neural networks on the task of knowledge distillation. 

\subsection{The General EV3 Framework}

An overview of EV3 is provided in Algorithm~\ref{alg:ev3}, which describes the single-objective version of the algorithm. EV3 introduces a protocol of three steps: explore, assess, and adapt, which are described as follows:

\paragraph{Explore} In the explore step, various strategies are proposed to update the model parameters. This is a highly exploratory phase where ideas are generated, much like brainstorming. The use of different data subsets, optimizers, and learning rates fall under this category.

\paragraph{Assess} The assess step is mainly about comparing the efficacy of the different updates proposed in the explore step. The use of i.i.d samples ensures a fair comparison. 

\paragraph{Adapt} The adapt step is the most decisive. It is where we decide whether changes need to be made to the model, including its parameters and architecture, or to modify our optimization strategy going forward.

\begin{algorithm}[t!]
\caption{\fullEVE} \label{alg:ev3}
\begin{algorithmic}[1]
\Require
\begin{itemize}[noitemsep,topsep=0pt,leftmargin=12pt]
    \item[]
    \item Initial model graph $\graph_1$ belonging to a graph family $\graphfamily$
    \item Initial model parameters $\wt_1$
    \item Labeled data distribution $\data$
    \item Evaluation method $\metric$
    \item Smooth losses $\losses$ that correlate with $\metric$
\end{itemize}
\State $\wt^s = \wt_1$
\For{$t = 1, \ldots$}
  \item[] \hspace{5mm} \Comment{\bf I. Explore: propose candidate updates using losses.}
  \State Sample (potentially biased) gradient batches $\gbatch$
  \State $\grads = \text{EXPLORE}(\graph_t, \wt_t, \losses, \gbatch)$
  \smallskip\smallskip
  \item[] \hspace{5mm} \Comment{\bf II. Assess: select the best updates using the evaluation method.}
  \State  $\widehat{\wt} = \text{ASSESS}(\graph_t, \wt_t, \grads, \metric, \data)$
  \smallskip\smallskip
  \item[] \hspace{5mm} \Comment{\bf III. Adapt: decide how to update model parameters and graph.}
  \State $\graph_{t+1}, \wt^s, \wt_{t+1} = \text{ADAPT}(\graph_t, \wt^s, \wt_t, \widehat{\wt}, \metric, \data)$
\EndFor

\end{algorithmic}
\end{algorithm}

\subsection{EV3 for Knowledge Distillation}
Knowledge distillation (KD)~\citep{hinton2015distilling} is a representative methods for model compression and acceleration, which effectively learns a small student model from a large teacher model.  
It has received increasing attention from the community~\citep{gou2021knowledge}. 
In this work, we are particularly interested in applying EV3 to KD because KD methods often do not rely on labeled data for training.  
EV3 can thus use labeled data for validation to mitigate the risk of overfitting. We describe the essential components and design choices of using EV3 for KD in the following.

\subsubsection{Disentangling Exploration and Evaluation}
One major benefit of using EV3 for KD is that we can easily disentangle exploration and evaluation to reduce the risk of overfitting. 
In the explore step of EV3, we propose updates to the student model by distilling knowledge from the teacher model without using the labels. 
In the assess and adapt steps, the student model is evaluated on the labeled training data using metrics of interest. 
Based on this evaluation, we can decide which action to take, such as updating the model parameters or changing the model topology. 
In this work, we use accuracy as the evaluation metric for the image classification task. 
The adapt step employs a heuristic that updates the model parameters only if the validation accuracy significantly surpasses that of the current model, with a significance level set at 0.05. If there are more than a few updates that do not meet this criterion of significance, the heuristic will instead opt to expand the model.

\subsubsection{Model Expansion}
One useful feature of EV3 is the ability to increase model capacity when updating the parameters alone does not lead to improvements in performance. 
In order to retain the model’s performance when expanding the model, we adopt the approach of network morphism \citep{chen2016net2net,wei2017modularized,elsken2018efficient}. The basic idea is to initialize the parameters of the larger neural network based on the weights of the smaller network so that they have exactly the same output for any given input. Formally, a network morphism is a \emph{function-preserving} operator $(T, \widetilde{w})$ that satisfies $N_w(x) = (TN)_{\widetilde{w}}(x)$ for every input $x \in X$, where $N$ represents a model such as neural networks. In other words, $N_w$ and $(TN)_{\widetilde{w}}$ represent the same function. This can be achieved by properly initializing new parameters $\widetilde{w}$ based on the current parameters $w$. This performance-preserving initialization averts the large cost of training all architectures from scratch.  

In this work, we implement network morphism operators to expand a popular architecture in computer vision, deep residual networks (ResNet)~\citep{he2016deep}. 
We design a network morphism expansion method that deepens the network by adding a ResNet block at each feature resolution. 

\subsubsection{Synergetic Training} 

Utilizing the components discussed above, we implement two variants of EV3 for KD. 

\paragraph{EV3} We implement a basic EV3 optimizer without extensive exploration, \ie, only one update is generated in the explore step. If the model’s performance does not improve significantly for a few iterations, we expand the model using a network morphism in the adapt step. In this work, the significance test is performed as a two-population z-score test with a confidence level of $\alpha=95\%$. 

\paragraph{EV3-Synergy}
Recent work by \citet{wang2023improving} indicates that KD methods do not always yield improved results when using larger models as the source of knowledge. Inspired by this finding, we introduce a synergetic training strategy for EV3, termed 'EV3-Synergy'. In this approach, the explore step involves proposing updates by distilling knowledge from a variety of models with different sizes. The process of EV3-Synergy begins by using the basic EV3 framework to generate a diverse set of models. We then further refine these models by running EV3 again, this time using both the initial teacher model and the newly created model collection as sources for distillation. The underlying rationale for this synergetic approach is the hypothesis that intermediate-sized models might lead to better performance outcomes compared to solely relying on the largest model for KD.

\section{Experiments}

To demonstrate the capability of EV3, we evaluate it on a knowledge distillation task for image classification on CIFAR-100~\citep{krizhevsky2009learning}. 
The teacher model is ViT-B/16~\citep{dosovitskiy2020image} and the students are ResNet models with different numbers of layers (ResNet-8/14/26/50).  
In terms of sizes, the teacher model has 86M parameters, and the student models vary from 0.08M to 0.77M parameters.  
The teacher model is pre-trained on ImageNet~\citep{deng2009imagenet} and fine-tuned on CIFAR-100.  

\subsection{Implementation Details}

We compare EV3 to two baselines: vanilla KD and network morphism. Experiments are single-trial with the same random seed. 
The vanilla KD baseline trains models of different sizes independently through KD for a fixed schedule of 110k iterations. 
For network morphism, we start from ResNet-8 and iteratively expand to larger models (ResNet-14/26/50), and train a fixed schedule of 110k iterations before each expansion. 
For EV3 and EV3-Synergy, instead of using a pre-defined schedule, the methods automatically determines when to expand the model. The main difference between EV3 and EV3-Synergy is that EV3 proposes only one update in the explore step by distilling ViT. On the other hand, EV3-Synergy utilizes a multi-pass training strategy. The first pass is identical to EV3, but in the second pass, multiple updates are proposed in the explore phase by distilling from the ViT model as well as all the ResNet models with the best performance.

\subsection{Results}

\begin{figure}[t]
    \centering
    \includegraphics[width=.325\linewidth]{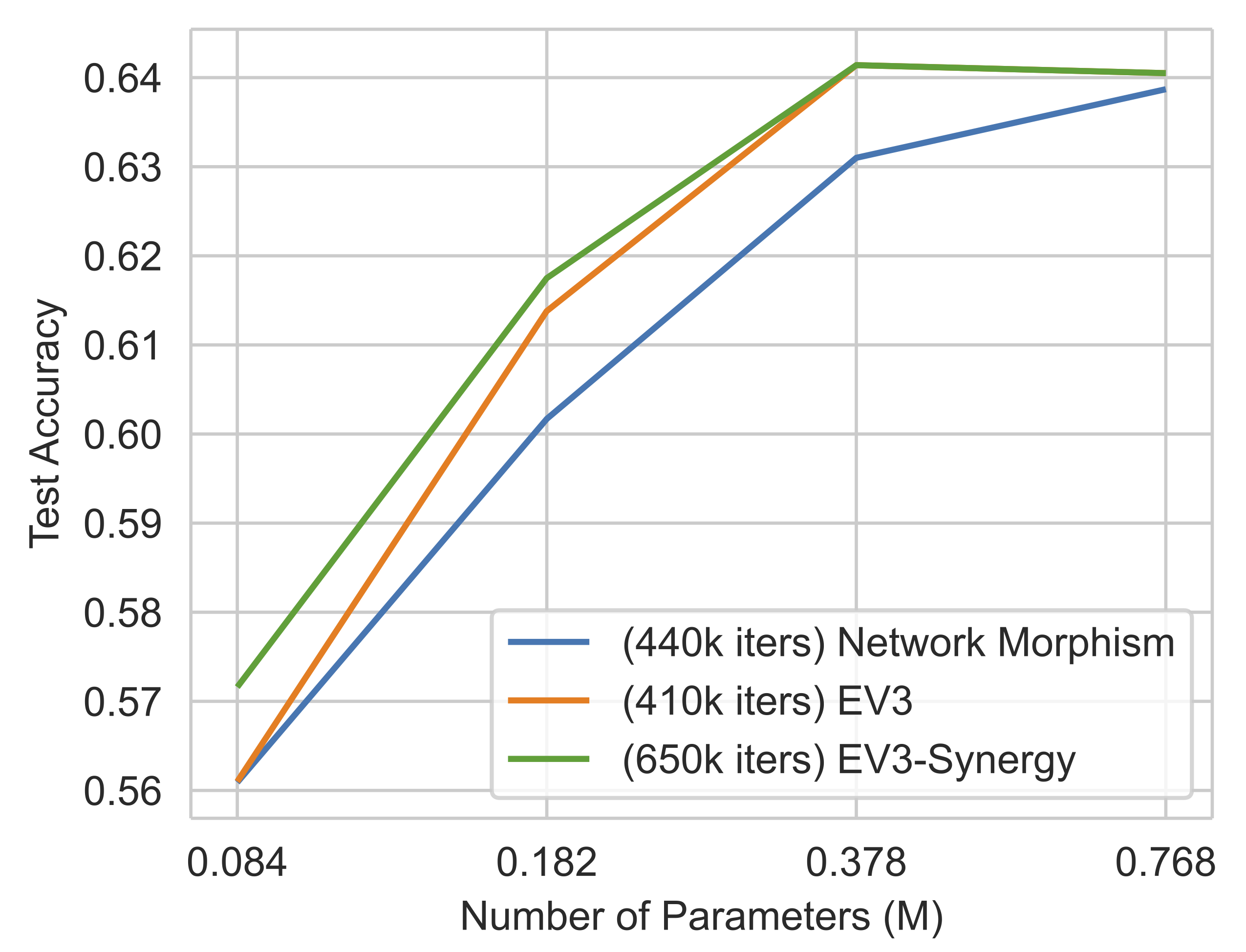}
    \includegraphics[width=.325\linewidth]{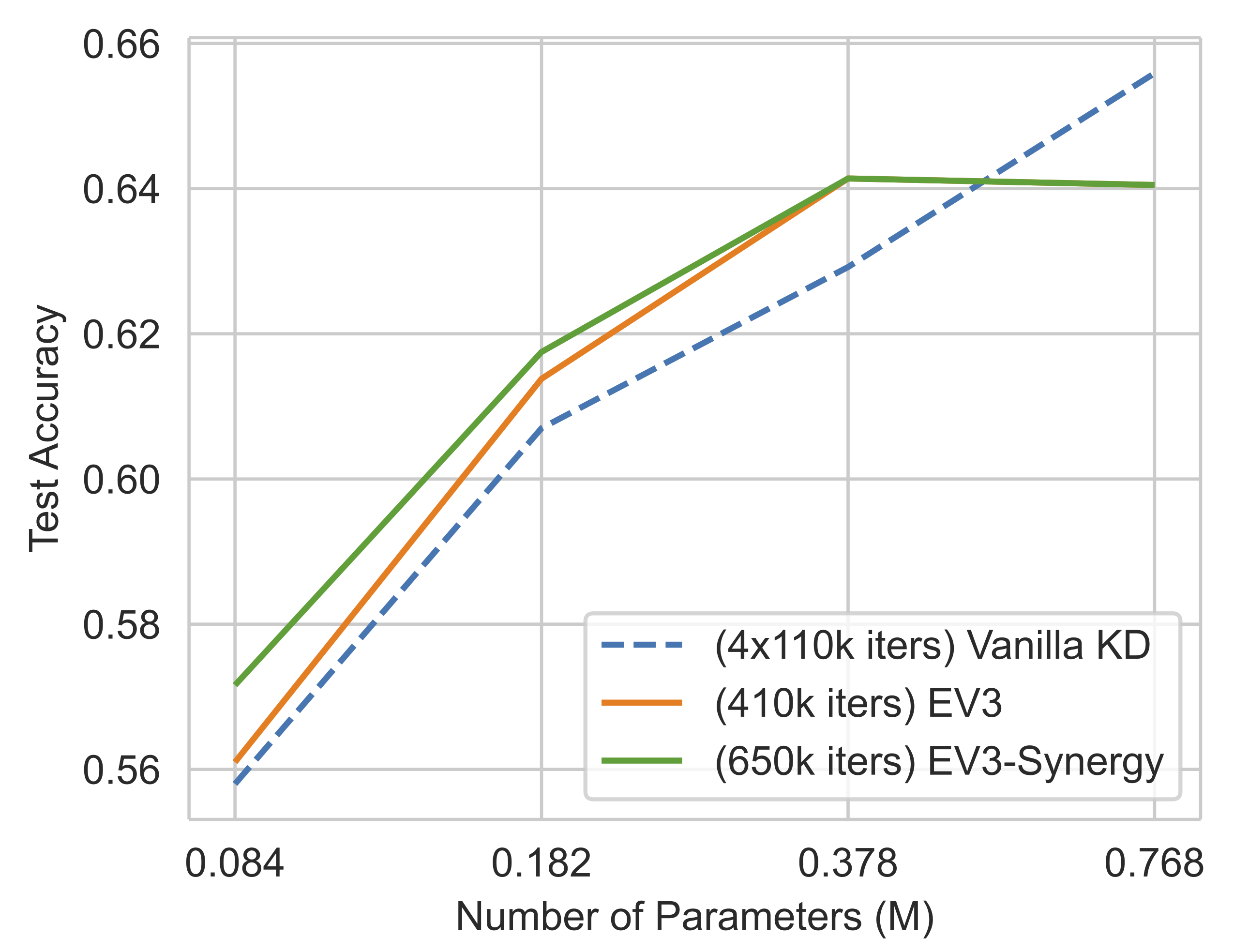}
    \includegraphics[width=.325\linewidth]{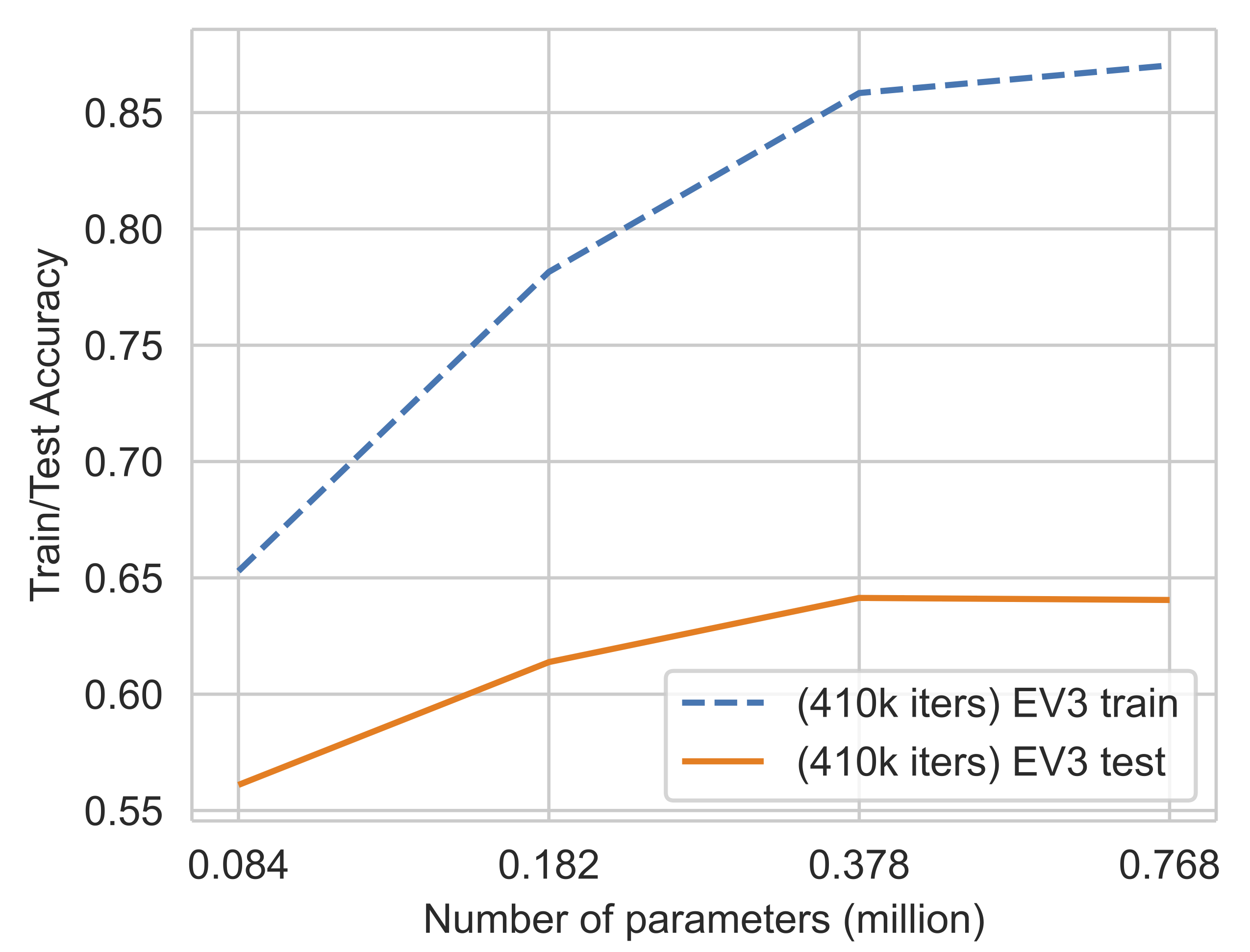}
    \vspace{-1mm}

    \caption{Results from left to right: 1) EV3 and EV3-Synergy compared to Network Morphism: Both EV3 and EV3-Synergy outperform Network Morphism. EV3-Synergy shows that the synergetic training strategy improves the performance of models of smaller sizes; 2) EV3 and EV3-Synergy compared to vanilla training: Both outperform vanilla KD except for the model of largest size. 3) Train vs. test error of EV3: the generalization gap becomes larger as the model size increases.}
    \label{fig:r1}
    
    \vspace{-7mm}
\end{figure}

The experimental results, as illustrated in Fig.\ref{fig:r1}, demonstrate the effectiveness of our proposed methods. Initially, in Fig.\ref{fig:r1} (left), we observe that EV3 outperforms network morphism in terms of accuracy, while maintaining similar computational costs. For instance, EV3 improves the accuracy of ResNet-14 (0.182M parameters) from 60.17 to 61.38, and ResNet-26 (0.378M parameters) from 63.10 to 64.14. This suggests that the adapt step of EV3 effectively finds a better schedule for model expansion using network morphism. Additionally, EV3-Synergy shows superior performance on smaller-scale models, such as enhancing the test accuracy of ResNet-8 from 56.10 to 57.16, validating our hypothesis that smaller models can benefit from distillation with intermediate-sized models.

The middle part of Fig.~\ref{fig:r1} indicates that EV3 generally outperforms the vanilla KD baseline across different model sizes, except for the largest one. Notably, neither EV3 nor network morphism surpasses the baseline on the largest model, suggesting two potential issues with model expansion via network morphism: 1) the possibility of being trapped in local optima due to multiple expansions, particularly when specific initializations (like identity weights in expanded layers) might deteriorate generalization, and 2) suboptimal scheduling for model expansion, potentially leading to overfitting, as indicated by EV3 selecting models with lower validation but higher test error.

Finally, Fig.~\ref{fig:r1} (right) displays the training error of EV3. It reveals an increasing generalization gap between training and test errors as the model size grows, suggesting a tendency of overfitting in larger models despite the absence of training labels in the EV3 updates. To address this in future work, we aim to conduct experiments with larger-scale datasets and explore the use of online data for KD.

\section{Related Work}

The proposed EV3 framework sits at the crossroads of several research areas. In this section, we discuss relevant research spanning evolutionary strategies, meta-gradients, neural architecture search, and preference-based reinforcement learning (RL).

\paragraph{Evolutionary Strategies}
Evolutionary algorithms (EAs)~\citep{back1993overview} have been an instrumental force in the optimization of non-differentiable functions, leveraging mechanisms inspired by biological evolution such as reproduction, mutation, and selection. Recent research has focused on using EAs in machine learning, especially deep networks~\citep{such2017deep,jaderberg2017population,ding2021evolving}. There has also been a trend of using hybrid methods of gradient descend and components of EA~\citep{cui2018evolutionary,ding2021optimizing,ding2022lexicase}. In a broader scope, \citet{real2020automl} proposed an evolutionary AutoML framework that automatically discovers complete algorithms using basic mathematical operations as building blocks. 
In line with current trends, we introduce a meta-optimization framework, which seeks to combine different learning approaches to provide a flexible tool for comprehensive training and development across various machine learning applications.

\paragraph{Meta-Learning}
Meta-learning aims to optimize the learning process itself and has seen substantial advancements, particularly in the domain of meta-gradients, which have emerged as pivotal for hyperparameter adaptation in non-stationary RL environments \citep{xu2018meta, luketina2022meta}. Research has delved into meta-optimizer characteristics, emphasizing the value of incorporating contextual information \citep{lange2023discovering,almeida2021generalizable}. Besides gradient-based approaches, black-box methods that parameterize the entire update rule as neural networks show promise \citep{andrychowicz2016learning, kirsch2022introducing}, while the intersection of bootstrapping techniques with meta-learning presents a nuanced dimension in RL scenarios \citep{flennerhag2021bootstrapped,guo2020bootstrap}. The EV3 framework, while encompassing aspects of meta-learning, offers a distinctive perspective that emphasizes an iterative exploration, assessment, and adaptation paradigm, as opposed to directly utilizing meta-gradients. It fundamentally diverges from traditional meta-gradient approaches due to its broad meta-optimization scope, weaving architectural evolution and task-specific metric optimizations, instead of solely focusing on adapting learning strategies.

\paragraph{Neural Architecture Search}
Neural Architecture Search (NAS)~\citep{elsken2019neural} focuses on automating the process of selecting the most suitable network architecture for a given task.  
Early methods in this area were computationally expensive, involving exhaustive search through a vast architecture space \citep{stanley2002evolving}.  
However, more recent approaches have employed efficient search strategies.  
\citet{real2017large}, \citet{xie2017genetic} and \citet{real2019regularized} showcased the use of evolutionary principles in NAS, while \citet{liu2018darts} proposed a gradient-based method to search the architecture space efficiently. In the context of our work, we mainly adopt the network morphism strategy~\citep{wei2017modularized,chen2016net2net,elsken2018efficient}, which aims to speed up training by initializing the weights of novel architectures based on weights of architectures that have been trained before. One critical challenge with network morphisms and other architecture expansion methods is the tuning of schedule for training and expansion, which we aim to solve by using the EV3 framework.

\paragraph{Preference-based Reinforcement Learning}
Traditional RL optimizes an objective defined by a scalar reward signal.  
However, Preference-based RL \citep{wirth2017survey} proposes a paradigm shift, where feedback is based on preferences between trajectories rather than scalar rewards.  
This form of feedback often aligns better with human judgment, especially when scalar rewards are hard to specify.  
Recent works like \citep{christiano2017deep,stiennon2020learning} have shown how such approaches can be deployed successfully in real-world scenarios, ensuring that AI systems capture human intent better.  
In the context of EV3, preference feedback can be used to guide the adapt step of the algorithm, illustrating our framework's flexibility and relevance to real-world optimization challenges.

\section{Conclusion}

In this paper, we propose EV3, a flexible meta-optimization framework designed for training machine learning (ML) models. 
Our investigation specifically focused on applying EV3 to knowledge distillation (KD) tasks. 
Experimental results demonstrate that EV3 outperforms network morphism in KD for image classification. 
Additionally, the EV3-Synergy variant, which incorporates a synergetic training strategy, further enhances these results.
Future work will aim to narrow down the generalization gap observed in EV3 by employing strategies such as dataset splitting and online training.
We also plan to extend EV3 to various other architectures, datasets, and tasks, such as using transformers for language-related tasks. 
Another potentially fruitful direction is the application of EV3 to multi-objective problems, including ML fairness.

\acks{%
We thank the Mudcats team, Esteban Real, and other colleagues at Google for helpful discussions, Lee Spector and Scott Niekum for their advice, and the anonymous reviewers for their thoughtful feedback.  
Moreover, the key ideas that led to the final design of EV3 benefited from discussions with many people, including Sebastian Bruch, Craig Boutilier, Suming J. Chen, Nando de Freitas, Harsh Mehta and Rich Zemel.  
We acknowledge the funding and computational resources provided by Google Research.
}

\vskip 0.2in
\bibliography{ev3}

\end{document}